\documentclass[]{article}

\usepackage{arxiv}

\usepackage[utf8]{inputenc} % allow utf-8 input
\usepackage[T1]{fontenc}    % use 8-bit T1 fonts

\usepackage{amsfonts}       % blackboard math symbols
\usepackage{nicefrac}       % compact symbols for 1/2, etc.
\usepackage{lipsum}		% Can be removed after putting your text content

\usepackage[square,sort,comma,numbers]{natbib}

\usepackage{doi}

\usepackage[onehalfspacing]{setspace}

\usepackage{booktabs}
\usepackage{tabularray}
\usepackage{tabularx}
\usepackage{siunitx} 
\usepackage{arydshln}

\usepackage{nicematrix}

\usepackage{graphicx}

\usepackage{hyperref}
\usepackage{url}
\usepackage{subfig}
\usepackage{caption}
\usepackage{subcaption}
\usepackage{float}
\usepackage{amsmath}
\usepackage{hyperref}

\usepackage{authblk}

\RequirePackage{amsthm}

\usepackage{graphicx}%
\usepackage{multirow}%
\usepackage{amsmath,amssymb,amsfonts}%

\usepackage{mathrsfs}%
\usepackage[title]{appendix}%

\usepackage{textcomp}%
\usepackage{manyfoot}%
\usepackage{booktabs}%
\usepackage{algorithm}%
\usepackage{algorithmicx}%
\usepackage{algpseudocode}%
\usepackage{listings}%

\usepackage{lmodern}
\usepackage{anyfontsize}

\usepackage[utf8]{inputenc}
\usepackage[english]{babel}
\usepackage[dvipsnames,svgnames,x11names]{xcolor}

\usepackage{changes}
\makeatletter
\AddToHook{cmd/added/before}{\def\Changes@AuthorColor{Green}}
\AddToHook{cmd/deleted/before}{\def\Changes@AuthorColor{red}}

\usepackage{newtxtext,newtxmath}
\author[1,2]{Zahra Sadeghi\thanks{Corresponding author: zahras@dal.ca}}
\author[1]{Evangelos Milios}
\author[1,2]{Frank Rudzicz}

\affil[1]{Faculty of Computer Science, Dalhousie University, Canada}
\affil[2]{Vector Institute for Artificial Intelligence, Canada}

\date{}  

\begin{document}

\title{Exploring the features used for summary evaluation by Human and GPT}

\maketitle

\begin{abstract}
Summary assessment involves evaluating how well a generated summary reflects the key ideas and meaning of the source text, requiring a deep understanding of the content. Large Language Models (LLMs) have been used to automate this process, acting as judges to evaluate summaries with respect to the original text. While previous research investigated the alignment between LLMs and Human responses, it is not yet well understood what properties or features are exploited by them when asked to evaluate based on a particular quality dimension, and there has not been much attention towards mapping between evaluation scores and metrics. In this paper, we address this issue and discover features aligned with Human and Generative Pre-trained Transformers (GPTs) responses by studying statistical and machine learning metrics. Furthermore, we show that instructing GPTs to employ metrics used by Human can improve their judgment and conforming them better with human responses.
\end{abstract}

\section{Introduction}

Large Language Models (LLMs) have significantly advanced Artificial Intelligence (AI) tools and transformed the user experience. Generative Artificial Intelligence, particularly through GPT models, has made significant strides in delivering high-quality responses. However, the underlying mechanisms that allow these models to comprehend inquiries and generate appropriate outputs remain inadequately understood. In this context, it is not only important to generate rigorous methods for evaluation of their responses, but also analyzing how these models interpret the given prompts. In an area of work, LLMs have been used to self-assess their responses. In fact, a growing area of research focuses on employing LLMs for evaluation.  Using LLM as judges refers to an innovative evaluation technique where LLMs are guided to assess textual content based on predefined objective criteria. In this approach, the models are explicitly instructed on the steps and criteria relevant to the task, enabling them to evaluate and rate texts systematically. The primary task involves scoring a given text according to a described criteria based on the LLM’s language understanding capabilities.  
Recent research on employing Large Language Models as evaluators has shown promising results in automating the assessment of textual content across various domains, such as content quality evaluation, and model-generated text assessment. Studies have demonstrated that when provided with clear and objective evaluation criteria, GPTs can produce ratings that align with human judgments reliably \cite{wang2023chatgpt}. The judgment often is based on criteria that captures aspects such as coherence, relevance, and consistency. Researchers have also explored prompt engineering techniques to improve the consistency and reliability of LLM-based evaluations using reasoning approaches such as chain of thought \cite{liu2023g}. Furthermore, comparative analyses between traditional human-based assessment and LLM-driven scoring highlight a gap between LLM and human scoring. Several lines of research have been previously undertaken including the dimensionality and attribute of judgment, prompting strategies, and benchmark and datasets useful for doing this research domain. The idea of using LLMs as evaluators, rather than just content generators, shifts the focus from linguistic generation to cognitive emulation. Judging involves evaluative reasoning, which is central to human cognition. While human rely on semantic reasoning and logical understanding for making decisions, LLMs lack this ability. LLMs such as GPT-based models, have achieved remarkable success in natural language generation across a wide range of tasks. However, despite their strong performance on many tasks, they struggle with functional reasoning, mathematical understanding, precise number generation, and semantic consistency in ways that differ significantly from human intuition. LLMs are fundamentally based on pattern recognition in textual space. They have been trained on vast corpora of text data, but do not inherently possess a formal understanding of mathematical concepts, like assessment criteria. Instead, LLMs learn to generate plausible-seeming text based on what they have previously seen in the training data, without truly understanding the underlying mathematical principles.

In this research, we propose investigating to understand the underlying mechanisms used by GPTs to judge. More specifically, we are interested in analyzing the GPT responses to uncover the features they employ to score textual contents. This research aims to provide information on what GPTs evaluate when instructing to use a certain attribute for judgment. This opens opportunities to analyze how GPTs think. We also explore the extent to which GPT's internal metrics used for judgment align with Human, and if we can enhance this alignment by incorporating metrics employed by Human.

\section{Related Works}

LLMs can be used as evaluators to judge the quality of text generated outputs by machine learning models. Developing LLMs as judges is particularly valuable in areas that require fast and accurate evaluation at scale \cite{zheng2023judging} —especially when dealing with novel content or domains where humans may lack expertise or require extensive training. Hence, one area of machine learning research is focused on adapting LLMs for evaluation. In this regard, Yu et al. applied supervised and direct preference optimization to tune LLMs for the task of judgment \cite{yu2025improve}. One common way LLMs have been used for evaluating texts is through scoring, where LLMs are employed for judging the quality of generated content. Trivedi et al. proposed a method in which LLMs generate rationales for their judgments, then employ those to improve their scoring ability over iterative rounds, showing that rationales help with calibration and alignment with human judgments \cite{trivedi2024self}. Chan et al. worked on an Reinforcement Learning based mechanism to fine tune the performance of LLMs and to improve judging quality \cite{chan2025j1}. Xu et al. fine tuned LLaMA model using data generated with GPT-4 to perform a scoring task to rate generated text, identify specific errors, and explain them \cite{xu2023instructscore}. Chen et al. used Multi-modal LLM (MLLM) as a judge which benefits from both visual and textual domains for evaluation. 
Since there is no ground truth to assess the correctness or quality of the generated answers, the standard approach for evaluating the reliability of LLM responses is to compare them with human judgments, measuring their alignment through correlation metrics. For this purpose, several authors studied the correlation of LLM scores with Human and they found through effective prompting and guidelines it is possible to increase the correlation of LLM ratings with Human\cite{zheng2023judging}. Yang Liu et al. uses GPT‑4 with chain‑of‑thought to better align automatic evaluation with human judgments for summarization\cite{liu2023g}. Mendonça et al. evaluated the use of LLMs for automated scoring of formative assessments and the performance of LLMs is compared to that of human graders to understand the alignment between automated and human scoring\cite{mendoncca2025evaluating}. There has been a growing interest in leveraging LLMs for reference-free evaluation, motivated by the need to produce more human-aligned quality judgments \cite{ke2022ctrleval} \cite{shu2023fusion} \cite{hu2024themis}.
A prominent area of related research direction seeks automatic metrics that correlate with human assessments. The common automatic metrics used are BLUE \cite{papineni2002bleu}, ROUGE \cite{lin2004rouge} and METEOR \cite{banerjee2005meteor} which have demonstrated poor correlation with human judgment \cite{sulem2018bleu}. Another issue with these metrics is that model outputs need to be compared with a reference text or ground truth. Some of the reference-free metrics such as MoverScore and BERTScore presented better correlation with human judgments. MoverScore, uses contextual embeddings with Earth Mover Distance, for evaluating semantic similarity in generated text \cite{zhao2019moverscore}. BERTScore is a score based on BERT architecture which finds embedding similarity between reference and generated text \cite{zhang2019bertscore}. To better align LLM evaluations with human judgment, several researchers fine tuned Transformer models on human annotated datasets that include human-generated ratings \cite{amplayo2022smart}, \cite{sellam2020bleurt}. In BARTScore, a pre-trained BART model is used to compute the log-likelihood of a set of generated and reference pairs to assess quality \cite{yuan2021bartscore}. Ho et al. evaluated how well LLMs can judge in comparison with humans and showed that LLMs outperform other metrics such as EM and F1 for Question Answering task \cite{ho2025llm}. However, reference-free metrics often rely on a dedicated evaluation model that must be trained on human-labeled data, making them more complex and task-specific than traditional metrics. Huang et al. compared fine‑tuned open source models to GPT‑4 in multiple tasks and found that while fine‑tuned judges can perform very well in domain/tasks they are trained on, they often underperform compared to GPT‑4 when generalizing in out-of-domain settings \cite{huang2025empirical}.

\section{Method}
\subsection{Problem definition}
While textual processing and understanding depend on cognitive skills, recent studies have demonstrated a direct relationship between level of textual complexity and human text understanding \cite{amendum2018does}. Human text understanding is a complex cognitive process that involves various stages of comprehension, reasoning, and interpretation. Humans use many cues by relying on different aspects of textual complexity, such as syntax, semantics, pragmatics, context, and world experience. In contrast, LLMs mainly depend on pattern recognition rather than than on an actual understanding of the meaning of instructions. They use a statistical approach drawing on the documents seen in similar contexts and generate texts which are most probable based on their prior training. 
LLM as a judge automates the process of evaluation and scoring, however, it is still a black-box problem and there is a lack of transparency regarding the features used for generating evaluative scores. Our objective is to find out the statistical features that align well with Human and LLM scores. This will help to provide concrete definition for fundamental metrics that Human/LLM apply for evaluating texts. In this research, we investigate two dimensions of \textit{relevance} and \textit{coherence} for evaluation of generated summaries. Here, we leverage statistical feature analysis to find the most likely attributes of text that affect LLMs responses. 
The general problem formulation for a scoring task is as follows. Given an input $x \in \mathcal{X}$ predict a score $y \in Y$ that reflects the quality of $x$ according to a quality dimension using $E$, where $E: X \to Y$ is an evaluator system.
Consequently, having a dataset ${x_i}_{i=1}^n$, where each sample $x_i$ has a model-predicted score $ \hat{y}_i \in \mathbb{R} $, and a human-evaluated score $ y_i \in \mathbb{R} $, the objective is to compute the correlation between the sets of scores \( \{\hat{y}_i\}_{i=1}^n \) and \( \{y_i\}_{i=1}^n \). This is typically done using a statistical correlation measure such as the Pearson correlation coefficient, Spearman’s rank correlation or Kandall's Tau:

\begin{equation}
\rho = \frac{\sum_{i=1}^{n} (\hat{y}_i - \bar{\hat{y}})(y_i - \bar{y})}
{\sqrt{\sum_{i=1}^{n} (\hat{y}_i - \bar{\hat{y}})^2} \cdot \sqrt{\sum_{i=1}^{n} (y_i - \bar{y})^2}}
\end{equation}

\begin{equation} 
\rho_s = 1 - \frac{6 \sum d_i^2}{n(n^2 - 1)}
\end{equation}

where $d_i$ is the difference between the ranks of $y_i$ and $\hat{y}_i$.

\begin{equation}
\tau = \frac{C - D}{\frac{1}{2}n(n - 1)}
\end{equation}

where $C$ is the number of concordant pairs (when the rank of both samples agree), and $D$ is the number of discordant pairs.
Pearson correlation detects linear relationships between two continuous variables, while Spearman and Kendall correlations measure monotonic relationships, which can be either linear or non-linear.
\subsection{Features and metrics}
In this research, our objective is to find the underlying features used to evaluate textual content based on relevance and coherence dimensions in an unsupervised manner. To this end, we propose studying a set of machine learning metrics that measure statistical features from various aspects of textual content and examine their correlation with both LLM and Human scores. More specifically, for each input $x_i$, we apply a set of metrics $\{M_1, M_2, \ldots\}$ to obtain $m_i$ and then compare  $\rho_{m_i,\hat{y}_i}$  with $\rho_{m_i,{y}_i}$.
Initially, we look at metrics that measure linguistic complexity features. For this purpose, we use readability indices to approximate the level of difficulty and complexity of a textual content (Table \ref{tab:tab1}).

\begin{table}[ht!]
\centering
\caption{Readability indices}
\begin{tabular}{|l|p{5cm}|}
\hline
\textbf{Index} & \textbf{Description} \\
\hline
Flesch Reading Ease & The Flesch Reading Ease is a readability metric designed to indicate how easy or difficult a passage of English text is to read.\\
Automated Readability Index (ARI) &  The Automated Readability Index (ARI) is a readability metric that estimates the U.S. grade level required to understand a given text. It's commonly used to assess how easy or difficult written material is to read. \\ 
Flesch–Kincaid Grade &  The Flesch–Kincaid Grade Level is a readability metric that translates a piece of English text into a U.S. school grade level — indicating the minimum education someone needs to understand the text. \\
Gunning Fog index & The Gunning Fog index estimates the years of formal education a person needs to understand the text on the first reading.\\
SMOG Index &  The SMOG Index (Simple Measure of Gobbledygook) is a readability formula that estimates the years of education a person needs to understand a piece of English writing. \\ 
Dale–Chall Readability score & Dale–Chall Readability score directly counts difficult words \\
Lexicon Count &  Lexicon count is the total number of words in a given text — excluding punctuation. It’s essentially the word count, but cleaned up so only actual words are included (not symbols or punctuation marks). \\
Difficult Words & Difficult words index counts the number of unfamiliar or difficult words. \\
Syllables Count & Syllables count index counts the total number of syllables. \\
\hline
\end{tabular}
\label{tab:tab1}
\end{table}

Moreover, we focus on information theory metrics such as \textit{entropy}, \textit{perplexity} and \textit{sparsity} to quantify uncertainty, confidence and predictability features. Lower entropy and perplexity are associated with higher predictability and confidence. We also evaluate the sparsity and diversity level by applying the Gini Index which measures the distribution inequality feature within a text. Higher Gini reflects with more diversity and lower sparsity. 
\begin{equation}
    H(X) = - \sum_{i=1}^{n} p_i \log_2(p_i)
\end{equation}
\begin{equation}
    \text{Perplexity}(S) = \exp\left( -\frac{1}{|S|} \sum_{i=1}^{|S|} \log P(s_i | s_1, ..., s_{i-1}) \right)
\end{equation}
\begin{equation}
    \text{Gini} = 1 - \sum_{i=1}^{n} p_i^2
\end{equation}

To estimate the probabilities for perplexity and entropy, we utilize statistical word occurrence based on a unigram (1-gram) model. This method is straightforward, interpretable and efficient for small texts. However, it fails to capture long-range dependencies and lacks a deeper understanding of semantics. To overcome this limitation, we define a new metric that incorporates contextual information. To achieve this, we define the \textit{conditional perplexity} metric to assess the predictability of the summary conditioned on the input source. This metric shares similarities with BARTScore \cite{yuan2021bartscore}.
In this measurement, we compute the probability of each token conditioned on all the previous tokens, as well as the input source $I$. 
Here, we leverage the Cross-Entropy loss function and apply the pre-trained Microsoft/phi-2 language model developed by Microsoft Research. This Transformer-based modeling can capture complex contextual relationship and long-term dependence between sequence of words.  
\begin{equation}
    \text{Conditional\_Perplexity}(S | I) = \exp\left( -\frac{1}{|S|} \sum_{i=1}^{|S|} \log P(s_i | I,s_1, ..., s_{i-1}) \right)
\end{equation}

By applying the conditional probability metric, we ensure that source text is considered when evaluating the summaries. In addition, we assess \textit{cosine similarity} to the embedding representation obtained from applying sentence transformers to determine the amount of association between source and generated summary. This approach is closely related to BERTScore \cite{zhang2019bertscore}.
Finally, to find out the importance of each feature, we conduct correlational analysis. 

\subsection{Experimental results}
For evaluation of the results, we use human performance as the reference ground truth, comparing all results against human ratings. In order to measure the effectiveness of each statistical feature, we measure its correlation with both human and LLM judgments. In this paper, we focus on OpenAI's Generative Pre-trained Transformers (GPTs), as they represent one of the most advanced and powerful class of LLMs developed to date. These models are trained on vast datasets with advanced fine-tuning techniques and often outperform many open-source alternatives and can approach human-level performance. Moreover, GPTs have shown promise as a reliable Natural Language Generation evaluator, especially for tasks that require overall text quality assessments \cite{wang2023chatgpt}\cite{mann2020language}. We use SummEval dataset, a standard dataset created for text summarization evaluation \cite{fabbri2021summeval}. It includes human annotated scores for multiple quality dimensions. 
\begin{figure}[htbp]
    \centering
    \subfloat[Relevance]{%
        \includegraphics[width=0.9\textwidth]{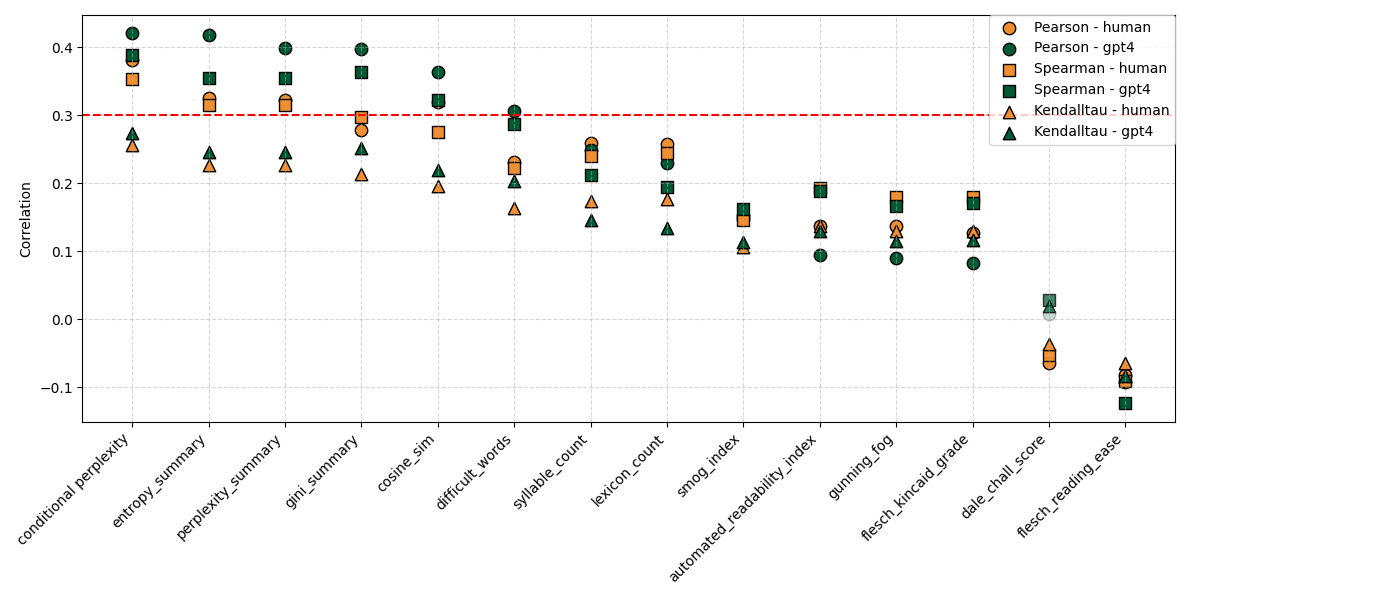}
        \label{fig:cor3-a}
    }\par\vspace{1em} % Line break + optional vertical space
    \subfloat[Coherence]{%
         \includegraphics[width=0.9\textwidth]{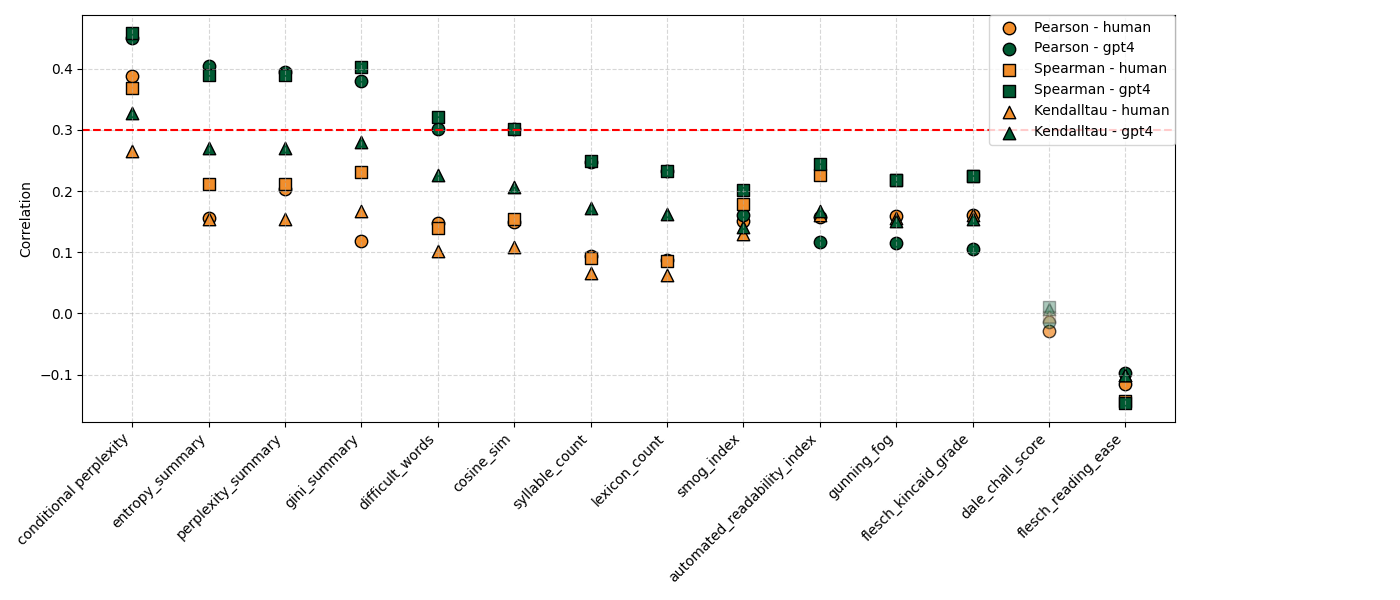}
        \label{fig:cor3-b}
    }
    \caption{Statistical features and their correlations. Color intensity reflects significance ($p\le 0.05$) ; greater intensity corresponds to greater significance. gpt4: GPT4-0613}
    \label{fig:cor3}
\end{figure}
The correlation between each metric and GPTs as well as Human are shown in Figures \ref{fig:cor3-a} and \ref{fig:cor3-b} for all three correlation types. Our results confirm that the proposed conditional perplexity metric maximizes the correlation with both Human and GPTs. The figures are sorted based on GPT4-0613 correlation with each of the features listed on the x-axis. We can observe that features based on uncertainty, predictability and diversity measured by Conditional perplexity, Entropy, Perplexity, and Gini index demonstrate the highest impact on GPT scoring. This confirms that GPT models, like other LLMs, rely heavily on statistical features driven from principles of information theory to understand text since they are essentially built on probabilistic models. In addition, it can be inferred that the complexity and difficulty of the text content measured by Difficult words, Syllable count and Lexicon count affect GPT responses. The results also suggest that Cosine similarity exhibits a superior alignment with the relevance dimension, which is affected by the proximity between source and summary in embedding space. 

\begin{figure}[htbp]
    \centering
    \subfloat[Relevance]{%
        \includegraphics[width=0.9\textwidth]{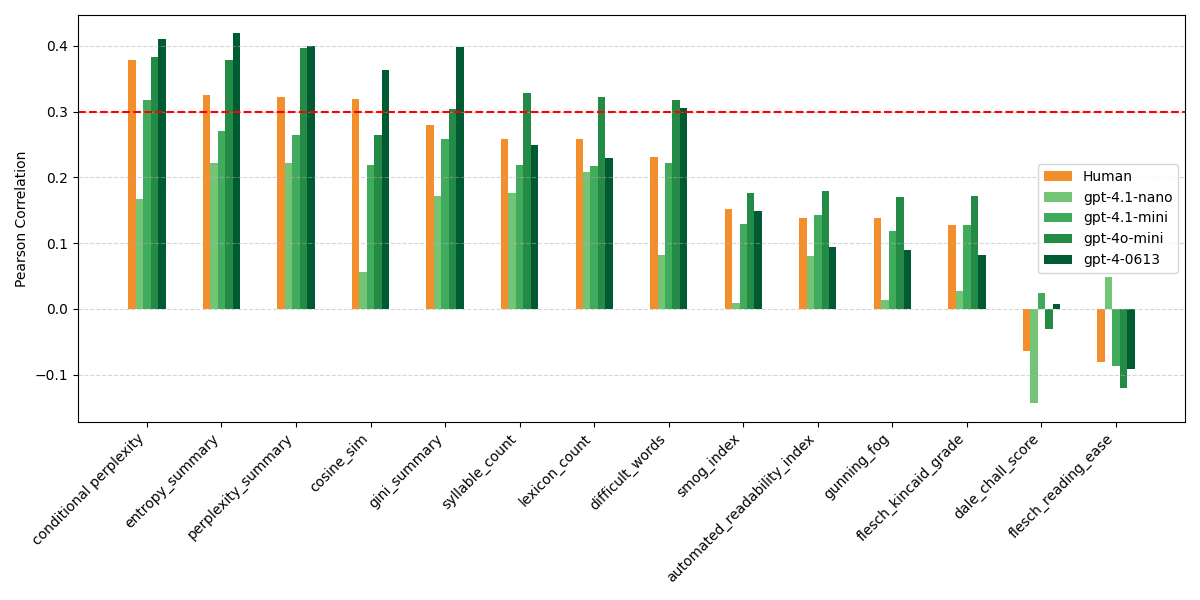}
        \label{fig:hgbar_rel}
    }\par\vspace{1em} % Line break + optional vertical space
    \subfloat[Coherence]{%
         \includegraphics[width=0.9\textwidth]{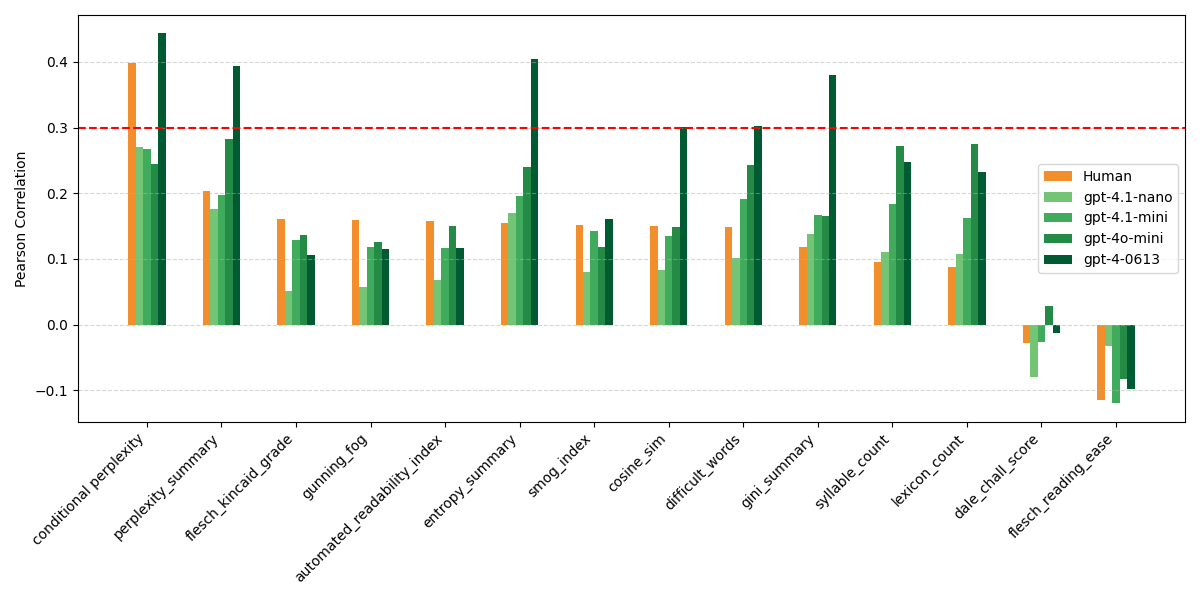}
        \label{fig:hgbar_coh}
    }
    \caption{Pearson correlations for all models and human}
    \label{fig:hgbar}
\end{figure}

\begin{figure}[hbt!]
    \centering
    \subfloat[Relevance]{%
        \includegraphics[width=0.7\textwidth]{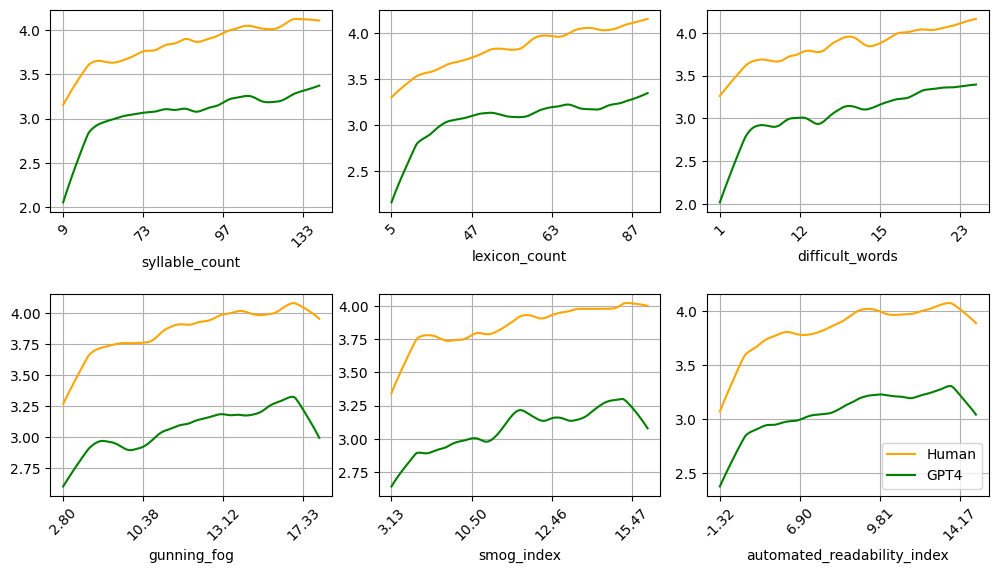}
        \label{fig:fig_low_rel}
    } 

    \subfloat[Coherence]{%
     \includegraphics[width=0.7\textwidth]{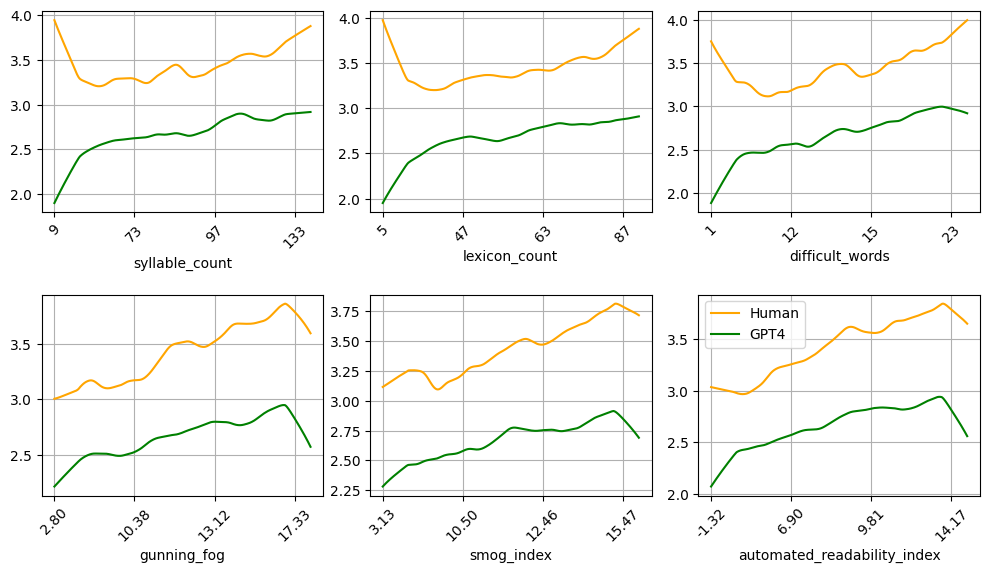}
        \label{fig:fig_low_coh}
    } 
    \caption{Readability indices vs Human/GPT evaluation scores }
    \label{fig:read_low}
\end{figure}

Moreover, to facilitate comparison, we provide bar charts for Pearson Correlation in Figures \ref{fig:hgbar_rel} and \ref{fig:hgbar_coh} sorted based on Human judgments for both the relevance and coherence dimensions. Similar to GPT's results, these figures indicate that the dimension of relevance generally exhibits higher correlations than coherence, suggesting that the studied metrics are more effective in identifying relevance between input text and summaries. In the same way, we observe the highest correlation achieved by the conditional Perplexity metric for Human scoring. The other top metrics contributing to relevance are Entropy, Perplexity, Cosine similarity, Gini index, Syllable count, Lexicon count and Difficult words. It can be concluded that the complexity of texts based on individual word characteristic quantified by lexicon count, syllables and difficult words influences Human scoring based on relevance. Other key features appear to be Perplexity, Flesch-Kincaid Grade, Gunnig Fog Index, Automated-Readability Index and Entropy. This suggests that coherence is primarily influenced by sentence-level linguistic features  as well as uncertainty based features within a text. This is noteworthy, as it emphasizes that in Human scoring, while word-level features play a critical role in relevance, coherence is primarily influenced by sentence-level features. In contrast, we observed that for GPT scoring, readability metrics relying on individual words achieved similar importance for both dimensions of relevance and coherence.   
To further analyze readability indices, we employ LOWESS method \cite{cleveland1979robust} as a non-parametric regression technique to fit a smooth curve to the human and GPT scoring ordered by the readability level for both dimensions of relevance and coherence in Figures \ref{fig:fig_low_rel} and \ref{fig:fig_low_coh}. We can observe that there is a nearly monotonic trend between readability and Human/GPT ratings for the relevance dimension. However, for coherence, there is a less obvious trend. Human tends to provide high ratings for the summaries with the lowest and highest levels of Syllable count, Lexicon count and Difficult words. We also observe a drop in Human and GPTs ratings for the summaries with the highest values of Gunning Fog Index, Smog Index, and Automated Readability Index.
Additionally, our results demonstrate that the lowest correlation is obtained using GPT-4.1-nano which is a compact model for light tasks. GPT4.1-mini which is a more efficient version, exhibits higher correlation. GPT-4o-mini supports a range of reasoning tasks and demonstrates better correlation. For GPT-4-0613 we use the ratings released by \cite{liu2023g}. Our experiments show that this version produces the highest correlation. In all experiments, we deploy the chain-of-thought prompting strategy introduced in \cite{liu2023g}. 

Furthermore, we investigate how to increase the correlation between GPT ratings with that of Human and we hypothesize that by instructing GPTs with the top correlated metrics with humans, we can encourage GPTs to follow a similar mechanism for evaluation. To this end, we add a new 'Hint' section, with which we augment the prompts by indicating the important metrics as shown in Table \ref{tab:tab2}. The results are summarized in Table \ref{tab:tab3} that confirms the effectiveness of this strategy.
\begin{table}[ht!]
\centering
\caption{Readability indices}
\begin{tabular}{|l|p{5cm}|}
\hline
\textbf{Dimension} & \textbf{Prompt augmentation} \\
\hline
Relevance &  Hint: \newline These metrics are important for evaluation:  \newline
- for summary: entropy, perplexity, Gini index, syllable count, lexicon count, difficult words
- cosine similarity between source and summary
 \\
\hline
Coherence &  Hint: \newline These metrics are important for evaluation of summary: perplexity, flesch kincaid grade, gunning fog, automated readability index, entropy, smog index \\

\hline
\end{tabular}

\label{tab:tab2}
\end{table}

\begin{table*}[ht]
\centering
\caption{Summary of correlations between Human and GPT ratings for relevance and coherence dimensions by applying 'hint' strategy}
\label{tab:geval-summary-correlations}
\begin{tabular}{lccccccc}
\toprule
 \multirow{2}{*}{\textbf{Evaluator}} & \multicolumn{3}{c}{\textbf{Relevance}} & \multicolumn{3}{c}{\textbf{Coherence}} & \\
\cmidrule(lr){2-4} \cmidrule(lr){5-7}
& Pearson $\rho$ & Spearman $\rho_s$
 & Kendall-Tau $\tau$ & Pearson $r$ & Spearman $\rho$ & Kendall-Tau $\tau$ \\
\midrule
GPT-4o-mini & 0.41& 0.39 & 0.30& 0.21 &0.20 & 0.16 \\
GPT-4o-mini + hint & \textbf{0.42} &  \textbf{0.41}  &  \textbf{0.31}  & \textbf{0.22} & \textbf{0.22} &  \textbf{0.18} \\

GPT-4.1-nano &0.20 & 0.20 & 0.15  &0.35 &0.36 & 0.28\\
GPT-4.1-nano + hint & \textbf{0.22} &  \textbf{0.21}  & \textbf{0.16} &\textbf{0.40 }  & \textbf{0.39} & \textbf{0.31}  \\

GPT-4.1-mini & 0.34& 0.34 &0.27 & 0.26 & 0.29 & 0.23 \\
GPT-4.1-mini + hint & 0.34 & \textbf{0.35}   &  \textbf{0.28}  & \textbf{0.32} & \textbf{0.34} &  \textbf{0.27} \\

\bottomrule
\end{tabular}
\label{tab:tab3}

\end{table*}

\section{Conclusion and future work}

 Semantic evaluation based on semantic dimensions like relevance and coherence is a complex task as there is no reference or ground truth to capture these conceptual qualities. However, the primary evaluation approach for Natural Language Generators (NLGs) such as LLMs, is based on comparing their precision against a reference. In this scenario, human ratings are the sole reference source. However, This approach has several limitations, as the ratings function like a black-box evaluation, offering no clear explanation of how the scores are determined and lacking any known metric they follow. In this paper, we investigated the features that explain the computation behind human and GPT ratings for the task of summary quality evaluation. Our findings suggest that both Human and GPT ratings are mostly influenced by uncertainty based features grounded in information theory, particularly through conditional perplexity metric, while they are also affected by textual complexity and readability features to varying degrees. We also showed that by augmenting the prompts with a hint about important metrics used by Human for each dimension, we can increase the correlation between GPT and Human scoring. In future work, more research needs to be conducted to find the exact internal calculation and thought process executed by black box models. In addition, further studies are necessary to design metrics that explain internal mechanisms performed by humans and LLMs in a closed mathematical format. Moreover, comparisons with human scoring may lack precision, as human raters can be influenced by factors such as fatigue, expertise in the subject, and the number and diversity of human evaluators in terms of their backgrounds and professions. In future research, greater attention and precision should be dedicated to collecting human data within a systematically controlled and rigorous environment.
\bibliographystyle{plain}  % or use IEEEtran, alpha, apalike, etc.
\bibliography{refs} 

\end{document}